# Accuracy Convergent Field Predictors

**Cristian Alb**

CA.PUBLICUS@GMAIL.COM

**Abstract**

Several predictive algorithms are described. Highlighted are variants that make predictions by superposing fields associated to the training data instances. They operate seamlessly with categorical, continuous, and mixed data. Predictive accuracy convergence is also discussed as a criteria for evaluating predictive algorithms. Methods are described on how to adapt algorithms in order to make them achieve predictive accuracy convergence.

## 1  Introduction

The family of predictive methods described in [1], also referred to as "concurrent data predictors" or "deodata predictors", is subdivided into several groups. Of interest are the subgroups referred to as "delanga" (proximity) and "rasturnat" (swapped). Note that, herein, the notion of predictor refers to a predictive algorithm.

A property that is shared by some variants of decision tree and deodata algorithms is discussed in [1]. The property in question is the predictive accuracy convergence. The importance of the property is argued herein. The deodata predictor family has many variants, but the convergence property has been shown to apply to only one of these. Other variants to which the property applies are detailed.

The swapped concurrent data predictors, or deodata rasturnat, are a subset of the deodata algorithms. They are important because they can be adapted to work with continuous attributes, not just categorical ones. They can be viewed as embodiments of field predictors in a feature space. It is shown that a particular variant can act as a bridge between two predictor types.

Also, the significance of singularity for the convergence of rasturnat algorithms is considered.

## 2  Deodata Predictor Types

Several types of predictive algorithm are detailed in [1]. A brief description of the functioning principles of the delanga and rasturnat variants follows. Note that in the description, a setting involving only categorical data is assumed.

Unlike decision trees [2], deodata predictors evaluate attributes concurrently and not in sequence. From a conceptual point of view, lazy learning, online algorithm, and instance-based learning are all descriptions that apply to deodata predictors. Implementations that are hybridizations with eager/offline algorithms are also possible. Such a hybrid scheme is suggested in [3].

A common characteristic of deodata predictors is the comparison of the attribute values of the training data entries with those of the query.

The training data set can be viewed as tabular data where each row corresponds to a training entry. Each column corresponds to an attribute/feature. A target outcome, also referred to as class or label, is associated to each training entry. The query can be viewed as a list of attribute values ordered in the sequence of attribute columns in the training data table.

The comparison with the query results in the computation of an entry match score for each row of the training data set. The entry match score is an aggregation of several column match scores. The column match score measures the degree of similarity between a pair of attribute values. The paired attributes correspond to the query and the training entry, respectively.

In the simplest case, the column match score can be a value of either one (1) or zero (0) depending on whether the attribute values match or not. Also, the aggregation of the column match scores can be a summation. In here, these choices are made and the entry match score represents the number of attribute values that match with the query. In this simple implementation, the entry match score is equivalent to a Hamming distance or overlap measure.

The way the entry match score is used differs among the deodata algorithm subfamilies.

It is to be noted that in this context, the attributes, or features, are considered to have the same degree of relevance for prediction. Adapting the algorithms to take into account the relevance of each feature is possible; the scores can be weighted with appropriate values.

## 3   Deodata Delanga (Proximity Concurrent Predictor)

The basic operation of the delanga variant consists in searching for the set of training entries that have the largest possible entry match score. The resulting set of entries will be used for prediction. The relative frequency of the outcomes in the predictive set provides an indication of their likelihood. For classification, a majority rule can be applied to the predictive set.

The operation is illustrated in Fig. 1.

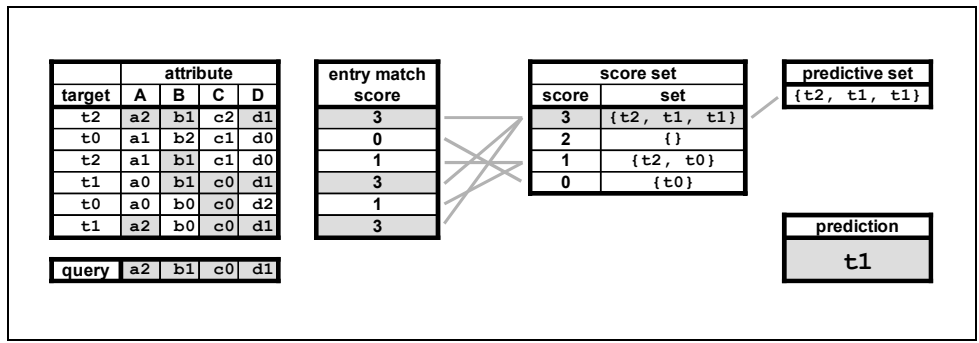

Figure 1: Deodata delanga operation example

A variation of the algorithm involves the addition of a tie-breaking mechanism. This is useful for classification tasks where the outcome with the highest number of occurrences in the predictive set has to be chosen. It can happen that there are count ties in the search for the most frequent outcome. The tie-breaking mechanism examines the runner-up predictive set and checks whether the count ties get broken there. Where the runner-up predictive set is the set of outcomes having the next best entry match score [1], [4].

## 4 Deodata Rasturnat (Swapped Concurrent Predictor)

In the rasturnat variant, the roles of the entry match scores and the outcome occurrences are swapped. The algorithm's operation consists in aggregating each entry's score to the corresponding outcome. The accumulated scores of each outcome provide a predictive measure of its likelihood.

The operation of the rasturnat variant involves an additional step, the evaluation of a transformation function. This function is applied to the entry match score and induces nonlinearity into the resulting entry transform score. This result is aggregated to the corresponding target outcome score.

The operation is illustrated in Fig. 2 where the transformation function is a power of two.

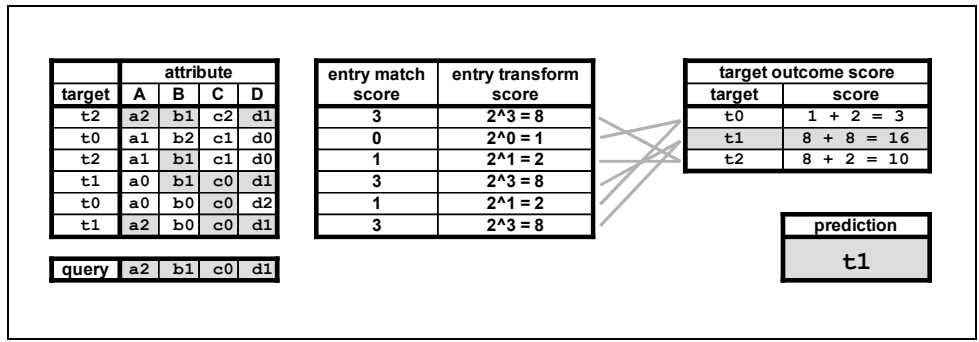

Figure 2: Deodata rasturnat operation example

Although other types of variations are possible, in here the focus is on rasturnat variations that are the result of modifying the transformation function.

An example of a simple transformation function is the exponential

$$f_{\text{TRANSF\_1}}(x) = 2^x \, , \qquad \text{(rasturnat\_pow\_2: a)}$$

$$f_{\text{TRANSF\_2}}(x) = e^x \, , \qquad \text{(rasturnat\_pow\_e: a)}$$

$$f_{\text{TRANSF\_3}}(x) = e^{x^2} \, . \qquad \text{(rasturnat\_gauss: a)}$$

Note that the entry match score is a measure of similarity between an entry and the query. A higher degree of similarity corresponds to a higher entry match score. A measure of dissimilarity can be defined as a complement to the entry match score. The complement can be the difference between the maximum possible entry match score minus the current entry match score. This dissimilarity measure is also referred to as matching distance.

For every function that takes as argument the entry match score, there exist an equivalent complementary function that takes as argument the matching distance.

For instance, with respect to the the previous examples,

$$f_{\text{CTRANS\_1}}(x) = \frac{1}{2^x} \, , \qquad \text{(rasturnat\_pow\_2: b)}$$

$$f_{\text{CTRANS\_2}}(x) = \frac{1}{e^x} ,$$ (rasturnat_pow_e: b)

$$f_{\text{CTRANS\_3}}(x) = e^{-x^2} .$$ (rasturnat_gauss: b)

The notion of kernel function is used herein to refer interchangeably to either the transformation function or the complementary function, depending on the context.

## 5  Predictive Accuracy Convergence

The predictive accuracy convergence is a property that characterizes the evolution of an algorithm's accuracy. It determines whether the prediction of the algorithm converges to an ideal accuracy as the amount of training data increases. An algorithm that has this property will be referred to as convergent.

The property, as presented in [1], applies to predictors and classifiers operating with categorical attributes and outcomes.

The task of comparing the accuracy of predictive algorithms is arduous because accuracy is influenced by data interdependencies specific to each problem setting. Presuming that algorithm *A* performs better than algorithm *B* for many problems, it is likely that there will be cases where the reverse happens. Performance rankings may change when algorithms are applied to a different problem setting. The convergence property offers a reference, or anchor point, in assessing and comparing predictive algorithms.

What about algorithms that deal with continuous or mixed attributes? In these situations the property is not directly applicable. However, if the continuous data predictor is an extension of a categorical algorithm, it could be argued that it has a kinship connection to the property. In such cases, the convergence in the categorical setting could be extrapolated to the continuous counterpart.

The deodata delanga predictor has the convergence property. This results from the following reasoning: as the number of available training entries increases, the probability of having training entries that match all the attributes of the query increases. As a result, the predictive set will correspond to a perfect entry match score. Also, as the number of training entries keeps increasing, the relative frequency of outcomes in the predictive set will converge toward the ideal distribution expected for the query's attribute combination. This is warranted by the law of large numbers.

## 6  Continuous Counterparts

So far the discussion has been centered on a setting involving data with categorical attributes. The scope of the algorithms can be expanded to include attributes taking continuous values. An intuitive way of achieving this is to replace the binary column match score with one that has a continuous degree of matching. Instead of having just a discrete set of possible values, one (1) for a match and zero (0) for a mismatch, the continuous [0, 1] interval can be used. The interpretation of the score remains consistent: one (1) for a perfect match and zero (0) for maximum dissimilarity. Such distances and similarity measures exist and can replace the column match score. The Gower distance is one such measure [5]. There is also the probabilistic distance as an alternative [6].

For the rasturnat variant, the substitution of the column match score is straightforward. Less obvious is the case of the delanga variant. The delanga algorithm uses the entry match score to establish a hierarchy of outcome predictive sets. A column match score that is continuous implies that the probability of having another entry with the same score tends to zero. As such, the set of outcomes with the best entry match score will most likely contain just one element. It follows that, in the continuous case, the delanga variant becomes equivalent to the 1-nn algorithm, a k-nearest neighbor with $k = 1$ [7]. This means that the algorithm always chooses as prediction the outcome of the nearest training entry.

## 7  Field Predictors

In a prediction task, the attributes of the training data can be thought of as defining a predictive space. The number of attributes corresponds to the dimensions of the space. The query, as every training entry, is just a point in this space.

Field predictor is a notion that describes an algorithm that bases its predictions on the influence of the training entries. The training entries are viewed as points creating a predictive field. The superposition of these fields determines the likelihood of the predicted outcomes. The rasturnat variant operates in this manner and constitutes an implementation of such a field predictor. However, the concept doesn't apply to the delanga variant.

An analogy can be made between a predictive field and the electric field. A setting can be considered where the predictive outcome is limited to two classes, defined as plus (+) and minus (-). The outcomes of the training entries can be thought of as electric point charges. Positive charges correspond to class plus (+) while negative charges correspond to class minus (-). The physical distance to an electric particle corresponds to the matching distance to a training entry. The predicted outcome for a query point in the attribute space can be assimilated to the sampling of the electric field in that same point. Such a setting is represented in Fig. 3.

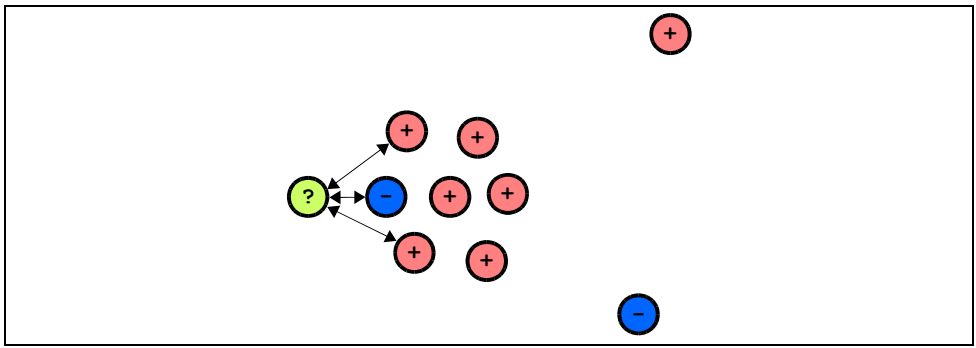

Figure 3: Electric field close sampling

In the figure, it is apparent that the electric potential at the sampling point must be negative. This is due to the close proximity of the negative particle. As a consequence, the outcome prediction corresponding to that query point must be minus (-). Both variants, delanga and rasturnat, are expected to make the same prediction.

The same type of setting is shown in Fig. 4, but with the notable difference that the sampling point is relatively far from the closest particle (negative). In this case, it is expected that the cumulative influence of the other particles' fields determines a positive potential. In this situation, the predictions of the delanga and rasturnat variants would differ.

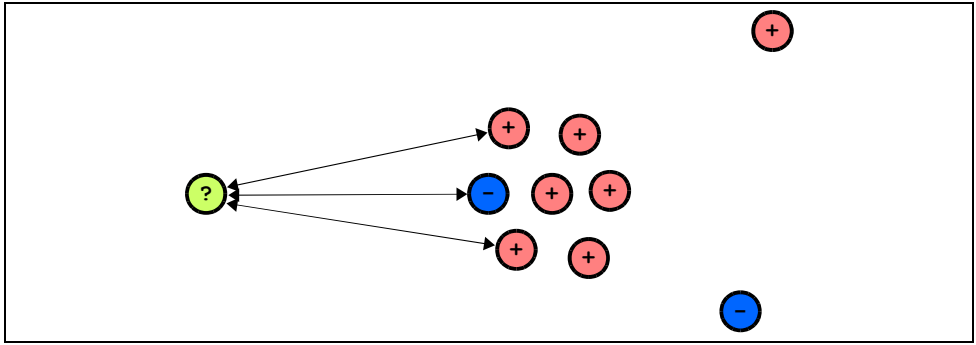

Figure 4: Electric field distant sampling

The delanga predictor looks exclusively at the nearest particles/entries. In this case, the closest particle is negative; therefore, the prediction would be minus (-). The rasturnat predictor cumulates the influence of all particles and, depending of the specific kernel function it uses, would most likely predict a plus (+).

It could be argued that the rasturnat variant is more in line with natural law, as in the electrostatic analogy.

## 8   Singularity and Convergence

Pursuing the field analogy with the realm of physics, it is notable the recurrence of the inverse-square law in the description of physical phenomena. For instance, it appears in the description of gravity and electrostatics. Inherent to the inverse-square law is singularity, the undefined behavior when the denominator is zero. As the argument of an inverse function approaches zero, the function's value increases and tends to infinity. For gravity, the singularity is associated to the concept of black hole. Albert Einstein is quoted as saying, "Black holes are where God divided by zero."

Does the importance of singularity for the gravitational field also translates into the domain of predictive field algorithms? Apparently, it does. A zero for the matching distance corresponds to a perfect match in the attribute space. The singularity corresponds to the largest possible entry match score. For the delanga predictor, a training entry that perfectly matches the query's attributes is added automatically to the maximally predictive set. For the rasturnat variant, the treatment of singularities is important. If processed appropriately, it can make a rasturnat variant achieve predictive accuracy convergence.

In this section, training entries matching all the attributes of the query are referred to as perfect match entries. Training entries having only one non matching attribute will be referred to as almost perfect match entries.

In order for a rasturnat variant to be convergent, it would suffice that each target outcome score be proportional to the number of training entries that perfectly match the query for that outcome. If this perfect proportion condition is fulfilled, for an ever increasing amount of training data, the target outcome scores will track the relative frequency of outcomes for that attribute combination. Again, this would be a consequence of the law of large numbers.

The perfect proportion condition requires that the entry transform score for perfect matches be substantially higher than that of entries with fewer matches. If the perfect entry transform score is guaranteed to be higher than the sum of entry transform scores of the remaining entries, the perfect proportion condition is fulfilled. The entry transform score for a perfect match needs to have an unassailable lead over the sum of the others. This would be possible if the transform score of almost perfect matches is a fraction of the perfect entry transform score. A fraction that can guarantee this unassailable lead is one over the total number of training entries.

Consider the following definitions:

$M_{\_TEN}$, the number of entries in the training data set;
$N_{\_TAC}$, the number of attribute columns in the training data set;
$TT[m, n]$, tabular training data matrix of size $M_{\_TEN}$ × $N_{\_TAC}$;
$q[n]$, query vector of size $N_{\_TAC}$;
$outc(i)$, the target outcome corresponding to row $i$ in the training data table $TT$;
$f_{\_TRANSF}()$, the score transformation function (kernel).

The following statements apply:

let $cms(q, i, j)$ be the column match score for attribute $j$ in row $i$ with respect to query $q$

$$cms(q, i, j) = f_{\_XOR}(q[j], TT[i,j]) ;$$

let $ems(q, i)$ be the entry match score for row $i$ with respect to query $q$

$$ems(q,i) = \sum_{j=0}^{N_{\_TAC}-1} cms(q,i,j) ;$$

let $ets(q, i)$ be the entry transform score for row $i$ in the training data table

$$ets(q, i) = f_{\_TRANSF}(ems(q, i)) ;$$

let $tos(q, k)$ be the target outcome score for outcome $k$ with respect to query $q$

$$tos(q,k) = \sum_{i:outc[i]=k} ets(q,i) ;$$

let $dm(q, i)$ be the matching distance to query $q$ from row $i$ in the training data table.

$$dm(q, i) = N_{\_TAC} - ems(q, i) ;$$

let $f_{\_CTRANS}()$ be the complementary function that takes as argument the matching distance (dissimilarity) instead of the entry match score

$$f_{\_CTRANS}(k) = f_{\_TRANSF}(N_{\_TAC} - k) ,$$

$$ets(q, i) = f_{\_TRANSF}(ems(q, i)) = f_{\_CTRANS}(dm(q, i)) .$$

Let's assume that the entry transform score is one (1.0) for a perfect match and it decreases monotonically as the number of attribute mismatches increases. Let the unassailable lead fraction be one over the total number of training entries. The most unfavorable case for convergence is when there is one training entry with a perfect match while all the other entries have almost perfect matches and are associated to another outcome. It is apparent that the sum of the entry transform scores of these imperfect matches will be less than one.

The following statements apply:

let *SEPM* be the perfect entry transform score

$$SEPM = f_{\_TRANSF}(N_{\_TAC}) \,,$$

$$\forall i \; \forall q \,, \; ems \, (q, i) \leq SEPM \,;$$

let *SEAP* be the almost perfect transform score corresponding to $(N_{\_TAC} - 1)$ attribute matches.

$$SEAP = f_{\_TRANSF}(N_{\_TAC} - 1) \,,$$

$$SEAP < SEPM \,;$$

let *MLD* be the multiplier lead of the perfect match score over the almost perfect one

$$MLD = SEPM / SEAP \,,$$

$$SEPM = MLD \cdot SEAP \,;$$

let *MAXSAP* be the maximum reminder summation of scores excepting one perfect match. This occurs when all reminder entries have almost perfect transform score.

$$MAXSAP = (M_{\_TEN} - 1) \cdot SEAP \,,$$

if $MLD = M_{\_TEN}$ :

$$\begin{aligned} MAXSAP &= (MLD - 1) \cdot SEAP \\ &< MLD \cdot SEAP \\ &< SEPM \,. \end{aligned}$$

As the number of training entries increases, more and more entries will match perfectly the query's attributes and the score weight of the less than perfect matches decreases.

As mentioned, the perfect proportion condition requires an unassailable lead for perfect matches. The transform score needs to be very large for zero mismatches. A parallel can be made with the huge intensity of a field in the vicinity of the singularity. Note that in the case of prediction, the transform scores should not be infinite. If they were, the values could not be added or compared.

## 9  Convergent Rasturnat Variants

Rasturnat variants that satisfy the predictive accuracy convergence requirement can be devised. Such algorithms are versatile; they satisfy both a theoretical requirement, the convergence, and a practical one, dealing with mixed data. It is possible to adapt an existing non-convergent algorithm into a convergent one. The solution consists in modifying the kernel function of the rasturnat predictor. Several methods are detailed in the following subsections:

- Splicing
- Scaled exponential
- Additive residuals
- Weakening exponential

Note that a categorical data setting is assumed in this section.

## 9.1 Splicing

This method could be applied to any existing rasturnat variant. It consists in substituting the kernel function value for perfect matches.
It takes the form

$$f_{\_TRANSF\_NEW}(k) = \begin{cases} f_{\_TRANSF\_OLD}(k) \text{, if } k < N_{\_TAC}, \\ \\ MLD \cdot f_{\_TRANSF\_OLD}(N_{\_TAC} - 1) \text{, if } k = N_{\_TAC}; \end{cases}$$

or equivalently

$$f_{\_CTRANS\_NEW}(k) = \begin{cases} f_{\_CTRANS\_OLD}(k) \text{, if } k > 0, \\ \\ MLD \cdot f_{\_CTRANS\_OLD}(1) \text{, if } k = 0; \end{cases}$$

where the MLD factor is chosen in such a manner as to insure the unassailable lead of perfect matches.
The disadvantage of this method is that it creates discontinuities. This would be apparent in a continuous data setting where the kernel function operates also with small fractional values.

## 9.2 Bridge Exponential

The inverse of an exponential function decreases monotonically if the base is greater than one. If the argument is the dissimilarity (matching distance) and the base is large enough, the function can act as the kernel of a convergent rasturnat variant.
For instance,

let $f_{\_CTRANS}(k) = 1/((b_{\_FACTOR})^k) = (b_{\_FACTOR})^{-k}$,

where $b_{\_FACTOR}$ is the base of the exponential function

$f_{\_CTRANS}(0) = b_{\_FACTOR} \cdot f_{\_CTRANS}(1)$.

More generally, for any $k$,

$$f_{\_CTRANS}(k) = b_{\_FACTOR} \cdot f_{\_CTRANS}(k+1) \ .$$

If $b_{\_FACTOR} = MLD$, the perfect proportion condition for convergence is satisfied. This results in

$$f_{\_CTRANS\_BRIDGE}(k) = 1/(MLD^{k}) = MLD^{-k} \ . \qquad \text{(rasturnat\_bridge)}$$

This variant has the property that, no matter what is the highest number of attribute matches, the champion entries will prevail in defining the predicted outcome. In the general case, the unassailable lead is guaranteed only for champion entries where all attributes match the query.

For classification, in case of a tie among the champion entries, the tie will be broken by entries with less attribute matches. This means that the bridge exponential operation is equivalent to the tie-break deodata delanga variant.

In general, rasturnat variants act as a bridge between continuous and categorical attributes. Additionally, this variant acts as a bridge between delanga and rasturnat algorithm types/subfamilies. Conceptually, this reconciles local methods (delanga) with global ones (rasturnat).

## 9.3    Inverse Additive Residue

Any rasturnat variant whose kernel function is equivalent to the inverse of a monotonically increasing function can be adapted and made convergent. The denominator is modified by adding a residual term. The following examples illustrate the method.

Let $f_{\_GROW}()$ be a monotonically increasing function and let

$$f_{\_CTRANS\_OLD}(k) = 1/(f_{\_GROW}(k)) \ .$$

The resulting adapted function is

$$f_{\_CTRANS\_NEW}(k) = 1/(ADREZ + f_{\_GROW}(k)) \ ,$$

where $ADREZ$ can be chosen as

$$ADREZ = (f_{\_GROW}(1)/MLD - f_{\_GROW}(0))/(1 - 1/MLD)$$

$$= \frac{\dfrac{f_{GROW}(1)}{MLD} - f_{GROW}(0)}{1 - \dfrac{1}{MLD}} \ .$$

As an exemplification,

$$f_{\_CTRANS\_OLD}(k) = 1/(2^{k}) \ ,$$

$$f_{\_CTRANS\_NEW}(k) = 1/(ADREZ + 2^{k}) \ ,$$

where

$$ADREZ = (2^1 / MLD - 2^0)/(1 - 1/MLD)$$

$$= -(M_{\_TEN} - 2)/(M_{\_TEN} - 1) .$$

This results in

$$f_{\_CTRANS\_NEW}(k) = 1/(2^k - (M_{\_TEN} - 2)/(M_{\_TEN} - 1))$$

$$= \frac{1}{2^k - \frac{(M_{TEN} - 2)}{(M_{TEN} - 1)}} .$$

(rasturnat_adj_pow_2)

A special case is represented by the Newtonian variant, an implementation of an inverse-square law field

$$f_{\_CTRANS\_NEWTON}(k) = 1/(1/M_{\_TEN} + k^2)$$

$$= \frac{1}{\frac{1}{M_{TEN}} + k^2} .$$

(rasturnat_newton)

## 9.4 Weakening Exponential

Other variants of interest are based on weakening exponential kernels. They are similar to the bridge exponential but with an exponent that does not increase linearly with the matching distance. The exponent rate of growth weakens, or fades, as the dissimilarity keeps increasing.

This mechanism is shown in the following examples

$$f_{\_CTRANS\_BRIDGE}(k) = 1/(MLD^k)$$

$$= 1/((MLD^1) \cdot (MLD^1) \cdot (MLD^1) \cdot (MLD^1) \cdot ... \cdot (MLD^1)) .$$

A fading variant is

$$f_{\_CTRANS\_DECAY\_A}(k) = 1/((MLD^{1/1}) \cdot (MLD^{1/2}) \cdot (MLD^{1/3}) \cdot (MLD^{1/4}) \cdot ... \cdot (MLD^{1/k})) ,$$

(deodata_rasturnat_decay_a)

or

$$f_{\_CTRANS\_DECAY\_B}(k) = 1/((MLD^{1/1}) \cdot (MLD^{1/4}) \cdot (MLD^{1/9}) \cdot (MLD^{1/16}) \cdot ... \cdot (MLD^{1/(k*k)})) .$$

(deodata_rasturnat_decay_b)

## 10  Field Sampling Density

The predictive field algorithms can be viewed as sampling a field in a point of the attribute space corresponding to the query's attributes. So far, it has been assumed that the entries of the training data are assimilated to electric point charges. What if, instead of electric charges, the entries are considered sampling points of the electric field? In this interpretation, an incongruity appears if the sampling is not done uniformly over the feature space. If most sampling is done in one small area, those field readings will have an excessive influence compared with readings scattered over larger areas. Such a situation is depicted in Fig. 4. In order to compensate this sampling density distortion, a weighting of the training entries can be applied. The compensation scheme should increase the score weight of entries in sparse areas while decreasing it in dense ones. In order to associate a density adjustment factor to each training entry, a self-similarity evaluation of the training data can be done as part of the training phase.

A possible density compensation scheme consists in:

- computing the compensation weights for the training entries into a data structure referred to as density compensation factor (*dcf*);

- modifying the prediction algorithm by inserting the additional step of multiplying each entry transform score with the corresponding density compensation factor.

The steps for computing the density compensation factor are the following:

- compute the virtual outcome score of each training entry with respect to all entries in the training data set. This value will be referred to as the entry's train similarity score (*tss*). In the context of this procedure, it is assumed that all training entries have only one possible virtual outcome

$$tss[j] = \sum_{i=0}^{M_{TEN}-1} ets(j,i) \; ;$$

- add together all computed scores into a sum of train scores (STS)

$$STS = \sum_{i=0}^{M_{TEN}-1} tss[i] \; ;$$

- divide the sum of train scores by the number of training entries. This results in the average train score (*STAVG*)

$$STAVG = STS / M_{\_TEN} \; ;$$

- compute for each entry the density compensation factor. It consists in the ratio between the average train score and the entry train score

$$dcf[j] = STAVG / tss[j] \; .$$

The density compensation factor can be thought of as an adjustment factor used in the calculation of the target outcome score. Alternatively, it can be viewed as a replacement of the standard entry transform score. After the usual computation of the entry transform score, the resulting value is multiplied with the entry's density compensation factor

$$ets_{STD}(q, i) = f_{TRANSF}(ems(q, i)) \,,$$

$$ets(q, i) = dcf[i] \cdot ets_{STD}(q, i) \,.$$

## 11 Conclusion

The importance of the predictive accuracy convergence property as a reference in evaluating algorithms has been discussed. Also, the significance of the deodata rasturnat predictors has been outlined. They implement prediction as a superposition of fields generated by the instances of the training data. They can operate seamlessly with both categorical and continuous data. Methods to adapt the deodata rasturnat algorithms in order to achieve predictive accuracy convergence have been presented. It has been shown how the concept of singularity plays a role in achieving this convergence. A convergent rasturnat variant has been highlighted for its bridging role between the rasturnat and delanga algorithm types. Additionally, a compensation method for sampling density distortions has been proposed.